\documentclass{ifacconf}

\usepackage{graphicx}      
\usepackage{natbib}        
  \usepackage{multirow}
    \usepackage{url}
    \usepackage{amssymb}
    \usepackage{amsmath}
\makeatletter
\DeclareRobustCommand{\pdot}{\mathbin{\mathpalette\pdot@\relax}}
\newcommand{\pdot@}[2]{%
	\ooalign{%
		$\m@th#1\circ$\cr
		\hidewidth$\m@th#1\cdot$\hidewidth\cr
	}%
}
\makeatother

\begin{document}
\begin{frontmatter}

\title{Deep Generative and Discriminative  Digital Twin endowed with  Variational Autoencoder for Unsupervised Predictive Thermal Condition  Monitoring of Physical  Robots in Industry 6.0 and Society 6.0} 


\author[First]{Eric Guiffo Kaigom} 

\address[First]{Department of Computer Science \& Engineering,\\
	Frankfurt University of Applied Sciences, Frankfurt a.M., Germany\\ (e-mail: kaigom@fra-uas.de).}


\begin{abstract}                
Robots are  unrelentingly used to achieve operational efficiency in Industry 4.0 along with symbiotic and sustainable  assistance for the work-force in  Industry 5.0. As resilience, robustness, and well-being are required in anti-fragile manufacturing and human-centric societal tasks, an autonomous anticipation and adaption to  thermal saturation and  burns due to motors overheating  become instrumental for   human safety and robot availability. Robots are thereby expected to self-sustain their performance and deliver user experience, in addition to communicating their capability to other agents in advance to ensure fully automated thermally feasible tasks, and prolong their lifetime without human intervention. However, the traditional robot shutdown, when facing an imminent thermal saturation, inhibits productivity in factories and comfort in the society, while cooling strategies are hard to implement after the robot acquisition. In this work, smart digital twins endowed with generative AI, i.e.,  variational autoencoders, are leveraged to manage thermally  anomalous and generate  uncritical robot states. The notion of thermal difficulty is derived from the reconstruction error of  variational autoencoders. A robot can use this score to predict, anticipate, and share the thermal feasibility of desired motion profiles  to meet  requirements from \mbox{emerging applications in Industry 6.0 and Society 6.0.} 

\end{abstract}

\begin{keyword}
  Robotics, Digital Twin, Gen AI, Unsupervised Machine Learning, Temperature Control,  Extended System Intelligence, Metarobotics, Variational Autoencoder, Deep Learning
\end{keyword}

\end{frontmatter}

\section{Introduction}
\begin{figure}[tp]
	\centerline{\includegraphics[width=0.67\columnwidth]{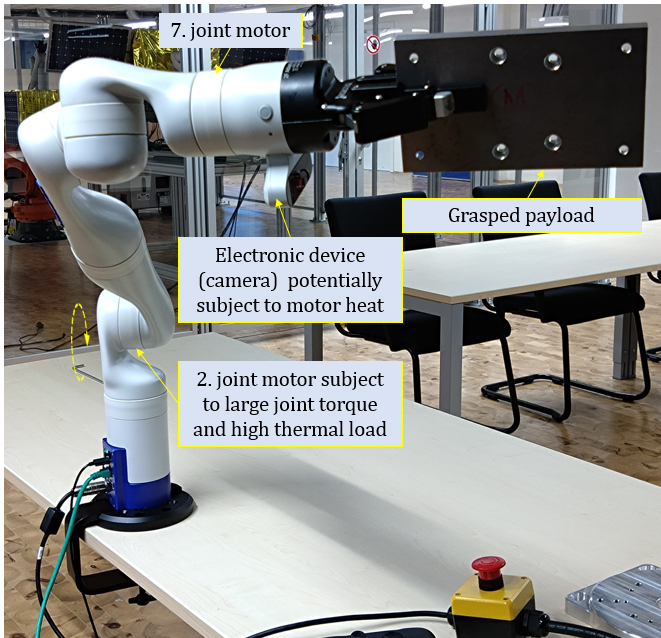}}
	\caption{Actively waiting robot under motor overheating.}
	\label{fig:sDT}
\end{figure}

Operating robots is often jeopardized by overheating  joint motors. Thermal burns due to motor overheating  is critical for the safety of humans in physical human-robot-interaction (\cite{Franka},\cite{Kinova}).  Thermal displacements of joints compromise  positioning accuracy and  affect  production consistency (\cite{soga2024skillful}). Thermal loads are continuously accentuated as robot arms maintain specific configurations (e.g., during active parking with a payload in-between manipulations) for a long period of time (see, e.g., the 2. joint in Fig. \ref{fig:sDT}). Over  years of operational usage, the motor temperature (e.g., around $130^{\circ}$C) (\cite{song2024overview}, \cite{zhu2021design}, \cite{singh2021thermal}) considerably impacts the degradation of the winding insulation. Included are the burnout of winding wire and deperming indicated in \cite{urata2008thermal}, as well as dielectric breakdown of winding insulation due to thermal aging (\cite{szamel2024monitoring}). Besides  limited peak torques impairing the robot performance (\cite{singh2021thermal}), thermal oxidation  might reduce the lifetime of electronic components (see Fig.~\ref{fig:sDT}), thus compromising  functionalities upon which the  {robot performance depends.} 

Strategies have been developed for the passive and active thermal management of joint motors. The active approach includes a forced convection that typically uses e.g. electric energy to accelerate  the transfer and quick accommodation of heat. By contrast, the passive strategy involves a natural conduction, convection, and radiation for heat absorption and dissipation. Passive cooling is known to be cost- and energy-effective as well as easier to implement than active cooling. However, active liquid cooling enhances the torque density (\cite{zhu2021design}, \cite{witzel2024analysis}). Practical insights in both thermal management techniques for robots are given in Table \ref{fig:table1} and \ref{fig:table2}. 
 Active immersive cooling of a BLDC motor of a legged robot has been shown to yield a thermal conductance that is $3.7$ times better than in the case of a forced liquid cooling, as shown in Table \ref{fig:table2}.


\begin{center}
	\begin{table}[t!]
			\caption{Thermal management for robots.
		}
	\begin{tabular}[Ht!]{ |p{2.1cm}||p{0.75cm}|p{2.2cm}|p{2.cm}|  }
		\hline
		\multicolumn{4}{|c|}{\textbf{Passive thermal management}} \\
		\hline
		\textbf{Strategy}& \textbf{Type} & \textbf{Thermal Conductivity (W/mK)} & \textbf{Application}\\
		\hline
		Heat sink (From gases to silver)  & Passive    &$0.01-10^3$(\cite{digikey})&   Electronic dev. \\ \hline
	    Heat spreader (e.g., diamond)  & Passive    &up to $2.2\cdot10^3$ (\cite{coherent})&  InP substrate,  LED, battery\\
		\hline
	\end{tabular}
\label{fig:table1}
\end{table}
\end{center}

\begin{center}
	\begin{table}[b!]
			\caption{Thermal management of a BLDC motor of a robot (based upon \cite{zhu2021design}).}
		\begin{tabular}[Ht!]{ |p{2.5cm}||p{2.3cm}|p{1.15cm}|p{1.15cm}|  }
			\hline
				\multicolumn{4}{|c|}{\textbf{Active thermal mngt of a motor of a legged robot}} \\
			\hline
			\textbf{Strategy (Room temperature $26^{\circ}$C)}& \textbf{Time $[s]$ to safety limit ($125^{\circ}$C)} & \textbf{Impro-vement in $[s]$} & \textbf{Impro-vement in $\%$}\\
\hline
\scriptsize{No Thermal Interface Materials (TIM) (Silicone)}&   3.6 &- &  -\\\hline
			\scriptsize{TIM + Forced Air (40 mm fan)} &   10.69  & $7$ &$197\%$\\\hline
			\scriptsize{Forced liquid (Water)} &11.59 &$7.9$&  $222\%$\\\hline
		\scriptsize{Immersion cooling (Deionized Water)}   &$\infty$ (i.e., Temp.   $\rightarrow 85^{\circ}$C $<125^{\circ}$C) & max. &  max.\\
			
			\hline
		\end{tabular}
		\label{fig:table2}
	\end{table}
\end{center}

Strategies mentioned thus far are relevant during the design and  manufacturing of robots. They are hard to apply once the robot is being operated, as pointed out in \cite{jorgensen2019thermal}, \cite{kawaharazuka2020estimation}, and \cite{singh2021thermal}. In this case, a software-based intelligent thermal control that harnesses joint states outputted by the application programming interface (API) becomes pivotal to anticipate thermal saturation. Manufacturing and even less emerging robot-driven sectors, such as healthcare$\slash$nursing and household, along with robotized space servicing (\cite{witzel2024analysis}), are missing such software. Deploying experts to actively manage thermal loads of  robots  can be prohibitive, as in space servicing (\cite{witzel2024analysis}). These sectors aim at using robots at a high level of availability, autonomy, and user experience. 

In this paper, we propose an unsupervised framework that predicts whether the desired motion of a robot  leads to overheating. The approach can generate thermally uncritical  motions to anticipate a thermal saturation. To this end, a digital twin (DTw) endowed with a variational autoencoder encodes (i.e., learns) the distribution of thermally uncritical states of the robot in a lower dimensional latent space. The notion of thermal difficulty derived from the decoding (i.e. reconstruction) error is introduced to predict thermal outlier motions and offer a measure that robots can share to communicate their risks of thermal saturation for a given time horizon. Our framework helps foresee and analyze thermal outlier states in both the latent and reconstruction space. It thereby offers not only visually rich and expressive but also numerical  insights in the thermal behavior of joint motors for the industry and society without thermal exposition or system disassembly. The AI$\slash$ML-embedded unsupervised approach targets a sustainable citizen well-being in Society 6.0 and collective intelligence in Industry 6.0 while following the non-invasiveness goals of our \mbox{overarching Metarobotics framework (\cite{kaigom2023metarobotics}).}

\begin{figure}[t!]
	\begin{center}
		\includegraphics[width=0.78\columnwidth]{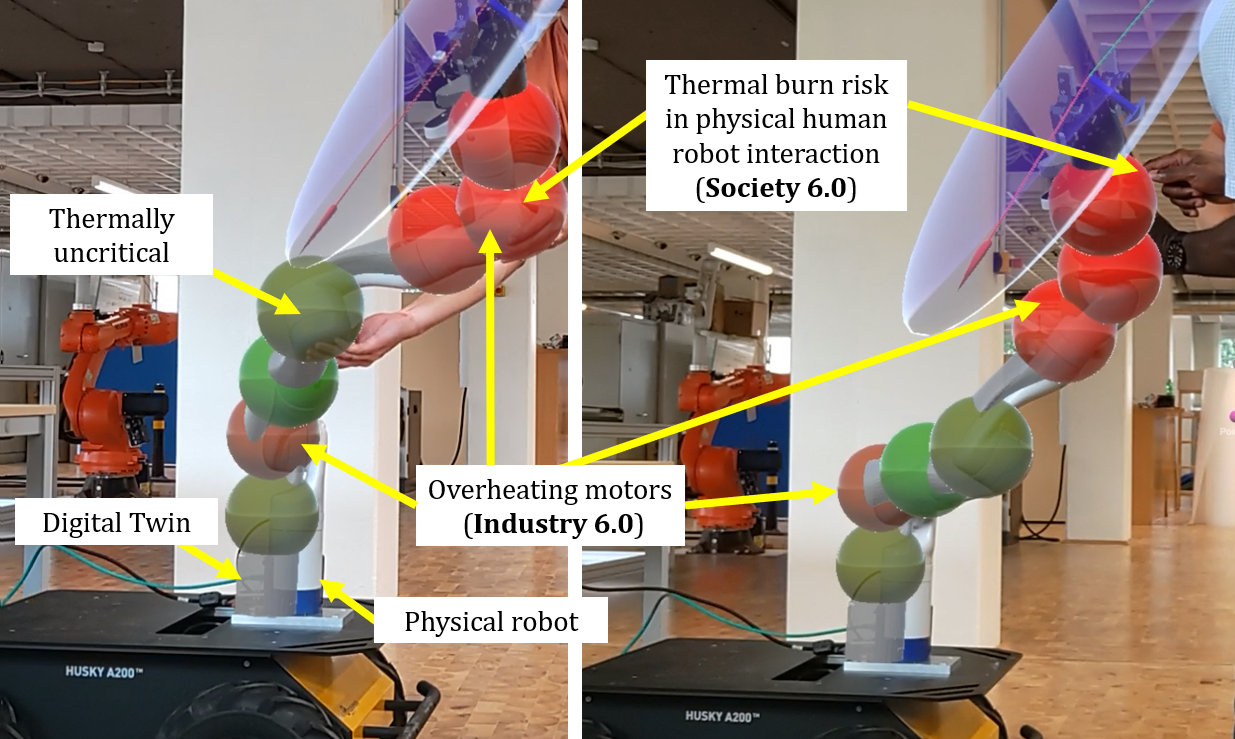}    
		\caption{Thermal monitoring (\cite{abt}) and issues. Sphere = thermal motor state. {Red/green = hot/cold.}} 
		\label{fig:i6}
	\end{center}
\end{figure}
\section{Related work} 
Efforts have been devoted to the management of the thermal behavior of robot joints. \cite{kawaharazuka2020estimation} leverage back-propagation to learn model parameters useful to estimate the core temperature of motors. The model can detect the augmentation of heat from the core to the housing that a motor of a musculoskeletal humanoid can face. In \cite{singh2021thermal},  thermal control takes advantage of peak current in robot actuators. A limiter is implemented to anticipate the output temperature of the stator that closely relates to the peak temperature at which the motor operates. An intelligent temperature controller for mobile robot is proposed by \cite{afaq2023intelligent}. Fuzzy logic is designed to actively regulate temperature with minimized energy demands. \cite{chen2024lifetime} predict  the lifetime of permanent magnet synchronous motors using the insulation thermal aging. The compensation of the thermal drift error for industrial robots is addressed in \cite{sigron2023compensation}. Difficulties to obtain temperature data of robots are pointed out as \textit{"One of the limiting factors..."} along with \textit{"...unfeasible system costs"} (\cite{sigron2023compensation}). In \cite{demirciouglu2023method}, an auto-encoder (AE) predicts thermal  anomalies of a DC motor through data compression and reconstruction. However, the distribution of data is omitted. The latent space is neither considered nor regularized. Discontinuities in the AE latent space are likely to  impair the  generation of  data.

By contrast, we learn in this work the  statistical distributions of observed  data about the state of  joints of a robot subject to a thermal process. The learned distribution is employed to detect motor overheating and create new thermally non-anomalous joint motions. 
We model the thermal dynamics as a generative process with stochastic hidden variables and assume Gaussian distributions. Unlike previous work, the loss function minimizes the reconstructions error and regularizes the latent space.
Variational autoencoding, as originally proposed in (\cite{kingma2017variational}), is the basic idea adopted in this work. Furthermore,  we introduce and explore the concept of \textit{thermal difficulty} derived from the reconstruction error of the variational autoencoder (VAE). This score offers a measure for the degree of outlierness of planned motions w.r.t the captured data distribution omitted in previous works mentioned thus far. We evaluate its connection to hyperparameters. VAE-driven DTw of robots can employ and share this numerical score for multiple purposes. Included are the  anticipation of thermal saturation and prevention of thermal burns in Industry 6.0 and Society 6.0 applications. We point out {data augmentation, autonomy, and parsimony of the VAE-based DTw.} We are not aware of any previous work that leverages VAE-embedded DTw to  monitor and anticipate the overheating of joints of an ultra-lightweight robot (see Fig.~\ref{fig:sDT}), to the best of our knowledge.

\section{Method}
\subsection{General formulation}
Assume the  dataset $X=\{x_1,\cdots,x_n\}$ is from a random thermal process. Each observed and multichannel
\begin{equation}\label{state}
	x_i=\begin{bmatrix}
		q_i \\\dot{q}_i \\\tau_i		
	\end{bmatrix}  \in \mathbb{R}^{3 \cdot n_J}
\end{equation}
 is a robot state.    $q_i$, $\dot{q}_i$, and $\tau_i$ are the joint positions,  velocities, and total {(including external)  torques at   timestamp $i$.} $n_J$ is the number of joints. For a redundant arm with $n_J=7$ as in Fig.~\ref{fig:sDT}, \mbox{for example, $x_i$ has the dimension $d_x=21$.}

Each $x=x_i\in X$ is associated with an unobserved and inducing latent variable $z=z_i \in Z=\{z_1,\cdots,z_n\}$ assumed to steer the process to produce $x$. 
According to Bayes, the \mbox{conditional probability distribution 	$p_{}(z|x)$ reads}
\begin{equation}
	p_{}(z|x)=\frac{p_{}(x|z)p_{}(z)}{p_{}(x)}.
\end{equation}
 The posterior distribution $p_{}(z|x)$ can be seen as encoding (e.g., projecting)  $x$ to yield a compressed lower dimensional $z$.  On the other hand,   $x$ is  reconstructed from the sampled $z$ with the decoding likelihood $p_{}(x|z)$, which needs to be maximized. 
Since the evidence  given by
\begin{equation}\label{bayes0}
	{p_{}(x)}=\int p_{}(x|z)p_{}(z)dz,
\end{equation}
 with prior $p_{}(z)$, is hard to compute for continuous $z$, the posterior $p_{}(z|x)$  is approximated  by a  {distribution $q_{\phi}(z| x)$  on the encoder side:}
\begin{equation}
	 q_{\phi}(z|x) \approx p_{}(z|x) .
\end{equation}
$\phi$ is a set of neural  parameters to be learned. The similarity between  $p_{}(z|x)$ and $q_{\phi}(z|x)$ is enforced by minimizing the Kullback-Leibler divergence $KL(q_{\phi}(z|x) || p_{}(z|x))$ of $q_{\phi}(z|x)$ from $p_{}(z|x)$. To this end,  the variational inference is leveraged. The minimization is equivalent to maximizing  
the evidence lower bound
\begin{align}\label{elboa}
	\mathcal{E} =& - 	KL(q_{\phi}(z|x) || p_{}(z)) \\\label{elbob}
	&+\mathbb{E}_{q_{\phi}(z|x)}[\log(p_{\theta}(x|z)]
\end{align}
as lower bound of $\log(p_{}(x))$ since $KL(a||b)\geq 0$ for any distribution $a$ and $b$. $\theta$ \mbox{are learnable neural network weights.}
\subsection{Specific training}
\begin{figure}[t!]
	\begin{center}
		\includegraphics[width=\linewidth]{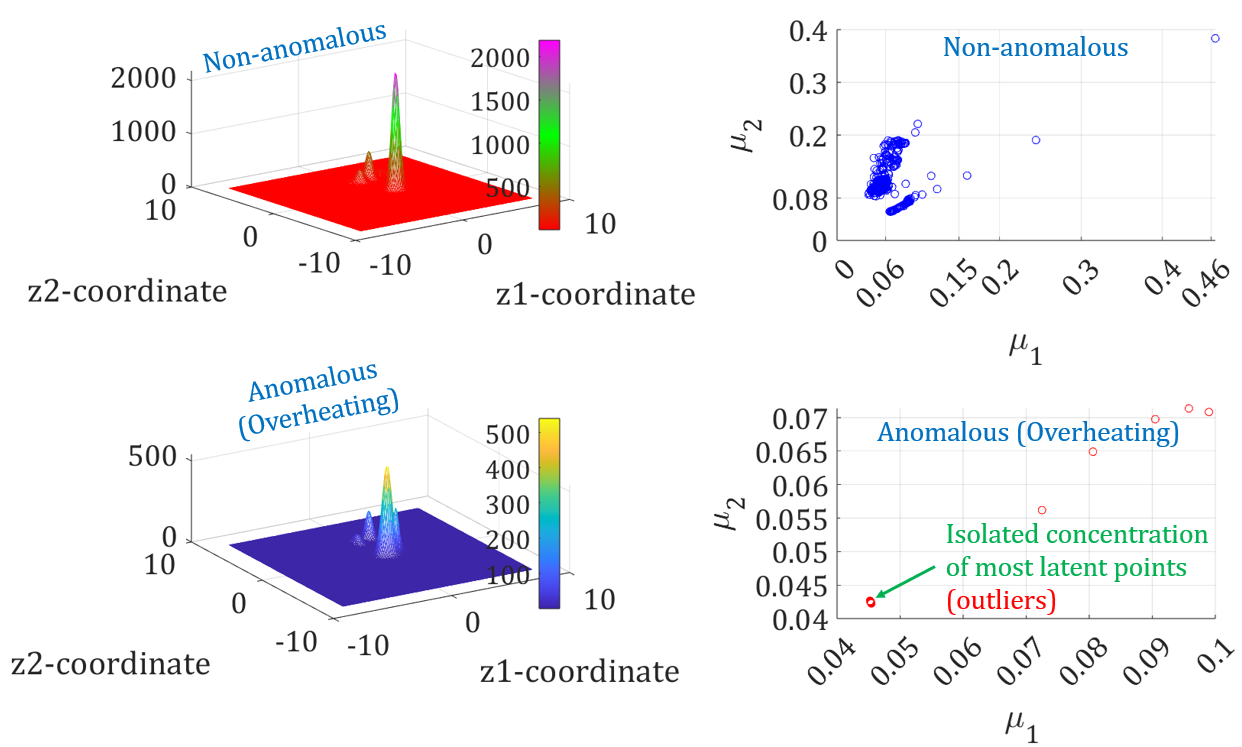}    
		\caption{L.h.s: VAE-based embedding of the  data manifold $(\text{the}~ x_i)$  into the  lower dimensional 2D latent space \textit{visualized} as superposition of learned Gaussian densities ($\mathcal{N}(\mu_{q_i},\Sigma_{q_i})$) related to $z_i$. R.h.s: $(\mu_1,\mu_2)$ associated with $q_{\phi}(z_i|x_i)$  for low$\slash$high temperature data in 2D. An outlier cluster  \mbox{is visible in the latent space.}} 
		\label{fig:latent}
	\end{center}
\end{figure}

In this work, the unknown latent vector has the form
\begin{equation}\label{latentz}
	z=	z_i=\begin{bmatrix}
		z_{1_i} \\ z_{2_i}
	\end{bmatrix}\in \mathbb{R}^{2}.
\end{equation}
Observe that $x$ in eq.~\ref{state} being compressed  has more components than $z$ in eq.~\ref{latentz} even for $n_J=1$. Latent variables and  their distributions assumed to be Gaussian can be easily visualized and further processed in the 2D latent space.  $ p_{}(z)$ is assumed to be a Gaussian distribution $\mathcal{N}(\mu_p,\Sigma_p)$ with the mean equal to $\mu_p = 0$ and the covariance matrix given by the identity $\Sigma_p=\begin{bmatrix} \sigma_{p_1}^2 & 0\\ 0 &\sigma_{p_2}^2 \end{bmatrix}= \begin{bmatrix} 1 & 0\\ 0 &1 \end{bmatrix}$. $q_{\phi}(z|x)$ is  viewed as a Gaussian distribution (see l.h.s of Fig. \ref{fig:latent}) $\mathcal{N}(\mu_q,\Sigma_q)$ whose parameters ($\mu_q, \Sigma_q$) with $\mu_q = \begin{bmatrix} \mu_{q_1} & \\\mu_{q_2} \end{bmatrix} $ and $\Sigma_q=\begin{bmatrix} \sigma_{q_1}^2 & 0\\ 0 &\sigma_{q_2}^2 \end{bmatrix}$ depends upon the weights $\phi$ of the deep neural encoding network to be trained. In this case, \mbox{the regularization term in eq.~\ref{elboa} is}
\begin{align*}
KL(q_{\phi}(z|x) || p_{}(z)) = -\frac{\beta}{2}\sum_{j=1}^{2} (1+\log(\sigma_{q_j}^2)-\mu_{q_j}^2-\sigma_{q_j}^2).
\end{align*}

The scalar $\beta \geq 1$  balances reconstruction and distribution capture, and fosters disentangled latents.   In the second term  (i.e., eq. \ref{elbob}), the conditional distribution is modeled as a Gaussian distribution, as in \cite{lin2019balancing}. This term  captures the expected reconstruction error  evaluated as the mean squared error between the output $x_{\text{pred.}}=\text{NN}_{\theta}(z)$ predicted by the  neural decoding network $\text{NN}_{\theta}$ (with input $z$ and weights  $\theta$) and the original input $x$ of the neural encoding network $\text{NN}_{\phi}(x)$ (see \cite{beaulac2019variational}).

Encoder and decoder neural networks made up of multiple long short-term memory (LSTM) and fully connected (FC) layers are trained using stochastic gradient descent and  backpropagation (\cite{Terbuch2022}). The reparameterization trick 
\begin{equation}\label{repara}
	 z=\mu + \epsilon \pdot \sigma
\end{equation}
is employed for a stable back-propagation. Indeed, eq.~\ref{repara} allows for the reduction of the variance estimate of the gradient when taking the derivative of the posterior distribution (\cite{ghojogh2021factor}). In eq. \ref{repara}, $\epsilon$ is a noise variable sampled on $\mathcal{N}(0,1)$. $\pdot$ is the Hadamar product used in eq. \ref{repara} to map  $\epsilon$ to $z$ \mbox{by taking advantage of $\mu$ and $\sigma$.}
 
\begin{figure}[t!]
	\begin{center}
		\includegraphics[width=0.9\columnwidth]{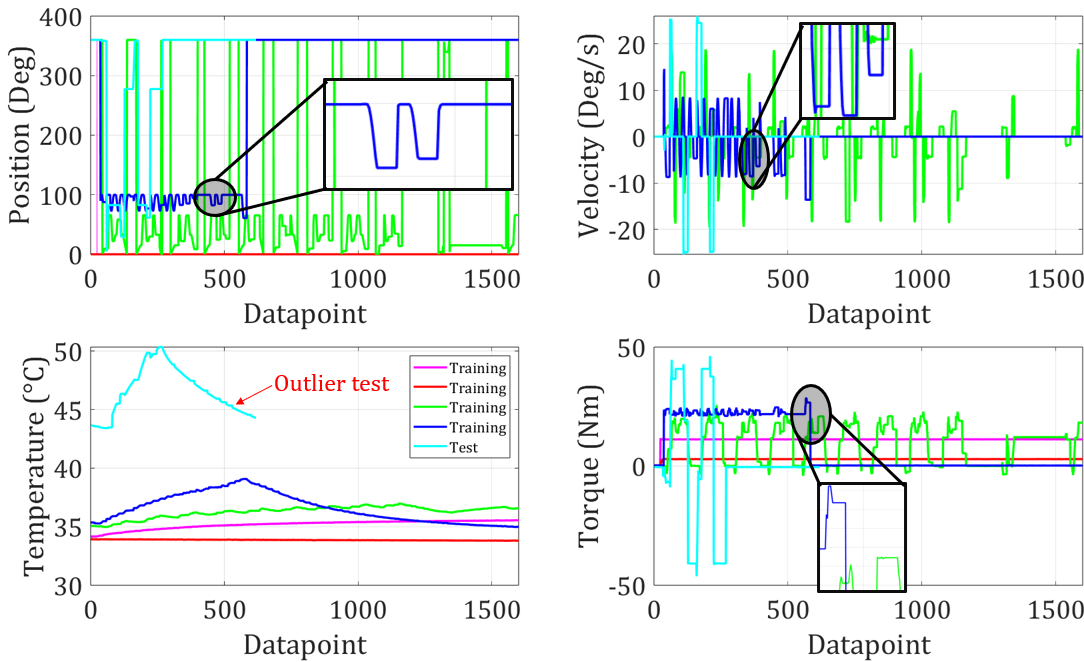}    
		\caption{Non-anomalous training ($q(t),\dot{q}(t),\tau(t)$) and cyan-colored anomalous test data with large temperature.} 
		\label{fig:training}
	\end{center}
\end{figure}
\section{Applications}
Experiments have been carried out to evaluate the usefulness of using VAE-endowed DTw to monitor the thermal behavior of the second joint of the physical robot depicted in Fig.~\ref{fig:sDT}. Furthermore, the capability of the VAE to generate thermally non-anomalous motions has been investigated. The DTw  endowed with VAE not only reflected the current state of the robot, but was also in charge of the aggregation and processing of datasets relevant for these objectives (Digital Shadow) while maintaining the traceability  of the usage of the physical robot (Digital Thread). Services of the DTw  leveraged the VAE to   detect outlier data out of the distribution from which thermally non-anomalous training datasets are for condition monitoring.

\begin{figure}[b!]
	\begin{center}
		\includegraphics[width=0.85\columnwidth]{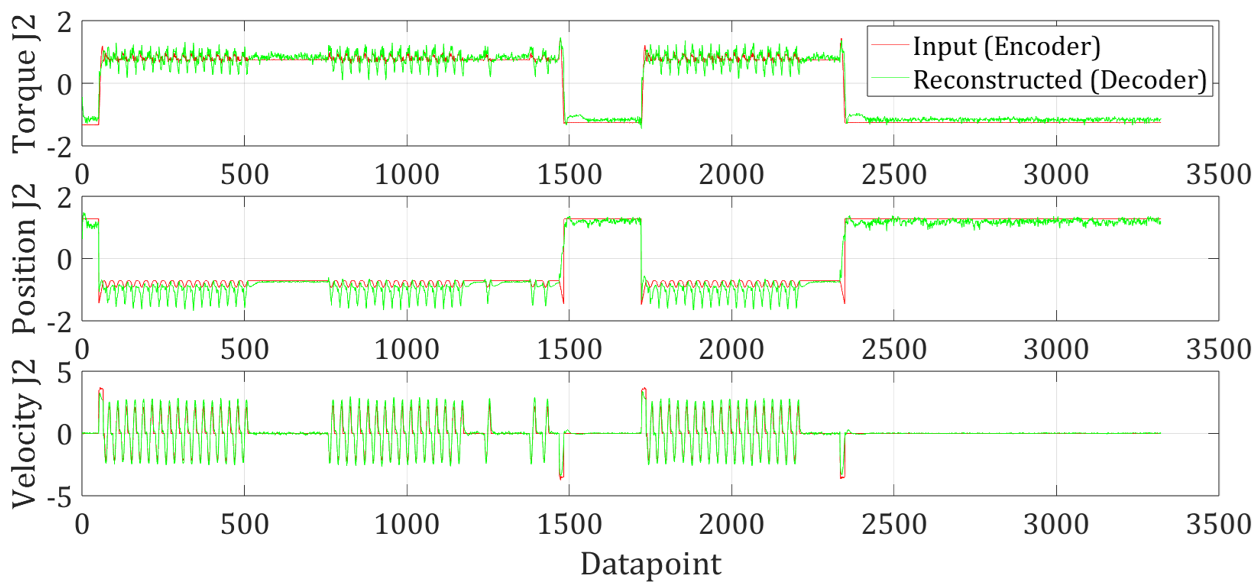}    
		\caption{Reconstruction of a non-anomalous dataset with the error smaller than in the critical case (Fig. \ref{fig:outluier}).} 
		\label{fig:rec}
	\end{center}
\end{figure}

\subsection{Data capture and pre-processing}
$q(t)$, $\dot{q}(t)$, and $\tau(t)$ were collected from the arm in Fig.~\ref{fig:sDT} to train the VAE. Examples of acquired datasets are shown in Fig.~\ref{fig:training}.   Datasets were obtained using  built-in sensors. Thermally uncritical (temperature below $40^{\circ}$) training and critical (temperature above $40^{\circ}$) test datasets were collected (see also Table \ref{tb:margins}). Each three channel training sequence had more than $1.5\cdot10^3$ datapoints (see  Fig.~\ref{fig:training}). The anomalous test dataset  had comparatively less data-points than training datasets and a higher temperature range (however below the warning threshold (see Table~\ref{tb:margins})), as shown in Fig.~\ref{fig:training}. The z-score  normalization  was used to pre-process the unfiltered training and test datasets. 

\begin{table}[h!]
	\begin{center}\label{thermaldata}
		\caption{Data for thermal  safety of the Kinova Gen 3 robot (see~\cite{Kinova})}\label{tb:margins}
		\begin{tabular}{ccccc}
			Data & Lower & Upper & Warning & Error \\\hline
			Max. motor temperature & $0^{\circ}$ & $80^{\circ}$ & $60^{\circ}$  & $75^{\circ}$ \\
			Max. core temperature & $0^{\circ}$ & $100^{\circ}$ & $80^{\circ}$ & $90^{\circ}$  \\ \hline
		\end{tabular}
	\end{center}
\end{table}

\subsection{Monitoring and anticipating thermal overheating}

\begin{figure}[b!]
	\begin{center}
		\includegraphics[width=0.9\columnwidth]{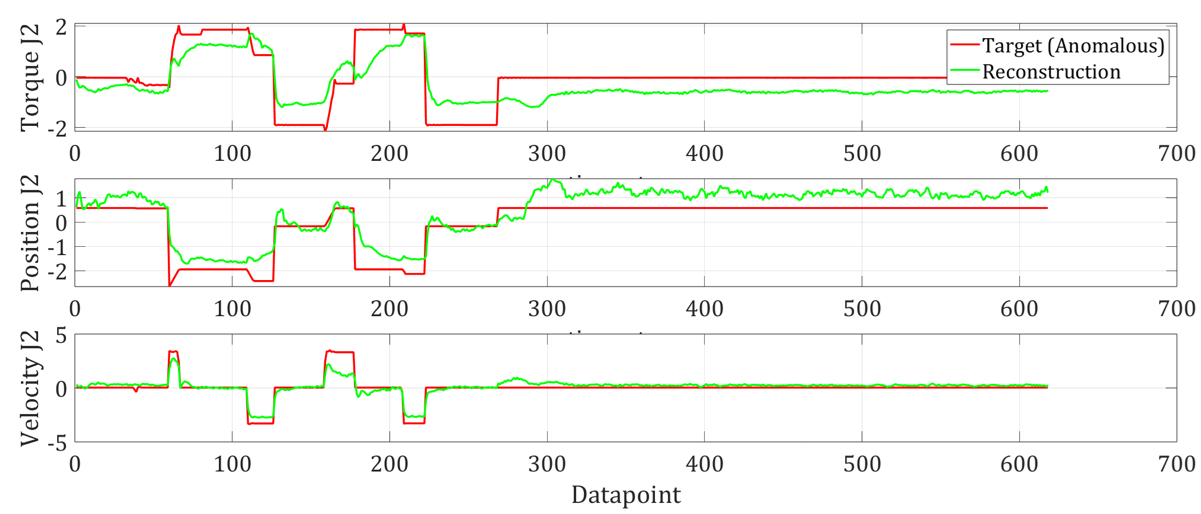}    
		\caption{Reconstruction of a high-temperature dataset. The  error is larger than in the non-critical case in Fig. \ref{fig:rec}.} 
		\label{fig:outluier}
	\end{center}
\end{figure}

The cumulation of Gaussian  probability distribution functions associated with the 2D latent vectors are depicted in Fig.~\ref{fig:latent}. Latent vectors related to non-anomalous training data are encoded in the upper part. Anomalous test data are projected in the lower part of Fig.~\ref{fig:latent}. Clusters of mean values can be observed in the lower r.h.s.  Distributions related to  data in the high temperature range (see lower l.h.s of Fig.~\ref{fig:training}) are apart from  remaining distributions, as indicated in the lower r.h.s of Fig.~\ref{fig:latent}. Sampling from distributions in this cluster generates motions \mbox{with a high overheating likelihood.}

\subsection{Detecting thermally anomalous motion profiles}

The goal was to detect joint states with high overheating potential by assessing issues encountered by the VAE to reconstruct original input data.  Anomalous datasets are characterized by a cumulative  reconstruction error between encoded and decoded data above a given threshold. The VAE was first trained using sensed data without any labeling. Hence, out-of-distribution datasets were  considered to alleviate over-fitting. Then, the  trained VAE was excited with  anomalous test datasets   from the  lower l.h.s of Fig.~\ref{fig:training}.  Fig.~\ref{fig:outluier} reveals that the VAE-DTw  hardly reconstructs original anomalous data in contrast to the non-anomalous case provided by Fig. \ref{fig:rec}. 
\subsection{Generating thermally non-anomalous motions }
Latent vectors were sampled according eq.~\ref{repara} during the decoding step. $z$ was obtained from  $\epsilon$ as stochastic input as well as  means and standard deviations of  approximated distributions related to thermally non-anomalous data. It was observed that generated states  are  close to  non-anomalous data for small perturbation in the latent space. Two reconstructions triggered by different  values of $\epsilon$ are shown Fig. \ref{fig:gen1}. Created position profiles are normalized and \mbox{thermally non-anomalous for the targeted joint motor.} 

\begin{figure*}[h!]
	\begin{center}
		\includegraphics[width=15.cm]{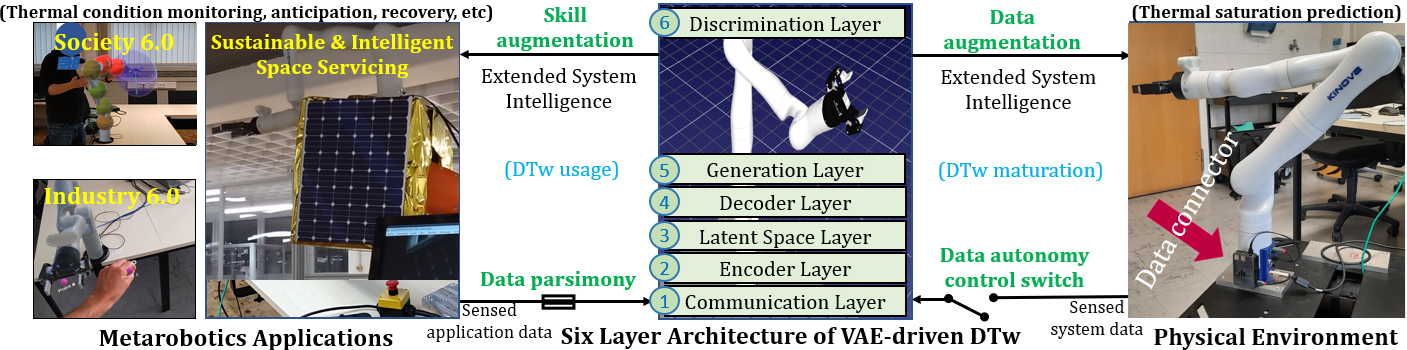}    
		\caption{Development, maturation, and usage of  VAE-based DTw for thermal  monitoring, analysis, and control in robotized applications. Data parsimony and  autonomy are  key \mbox{features of the underlying extended system intelligence.}} 
		\label{fig:outlook}
	\end{center}
\end{figure*}

\section{Discussion}
Technological, societal, and industrial implications of the  framework are discussed. Training insights are provided.
\subsection{Implications for the Digital Twin Concept}
A DTw aware of $p_{\theta}(x|z)$ and $ q_{\phi}(z|x) $ is advantageous for at least two reasons. First, the DTw can harness the encoder driven by $ q_{\phi}(z|x) $ to autonomously infer knowledge from datasets. Included are dimension reduction and clustering for data analysis and anomaly detection. Secondly, the DTw can use noise (i.e., $\epsilon$) to yield $z$ as input of the decoder driven by $p_{\theta}(x|z)$ to create new and similar (in terms of distribution) or augmented data for different  purposes.  Training  machine learning models that support services of the DTw on-the-fly is an example. After training, running the VAE-based DTw no longer necessitates a tightly coupled data flow from the physical hardware to the DTw. In fact, the DTw can generate its own datasets with relevance. This leads to a data parsimony and even autonomy of the DTw once it has built its stochastic generative model, as highlighted in Fig. \ref{fig:outlook}. In this case, the VAE model can serve as a basis for deep transfer learning to downstream applications. The idea is to employ a VAE-driven sustainable DTw to yield a  regularized and shared  latent feature space that fosters transferability and preserve discriminability across applications (see, e.g., \cite{ye2022learning}) while requiring less data (see  Fig. \ref{fig:outlook}).


\subsection{Collective intelligence implications  for Industry 6.0}
A robot endowed with a VAE-based DTw for thermal management can predict its thermal difficulty 
\begin{equation}
d=	\sum_{k=1}^{n_J} d_k
\end{equation}
that arises from a planned motion. Herein, 
\begin{equation}\label{TT}
	d_k=1-e^{-\frac{1}{n}\int_{t_l}^{t_h}|e_{r_k}(t)|dt}
\end{equation}

\begin{figure}[b!]
	\begin{center}
		\includegraphics[width=0.76\columnwidth]{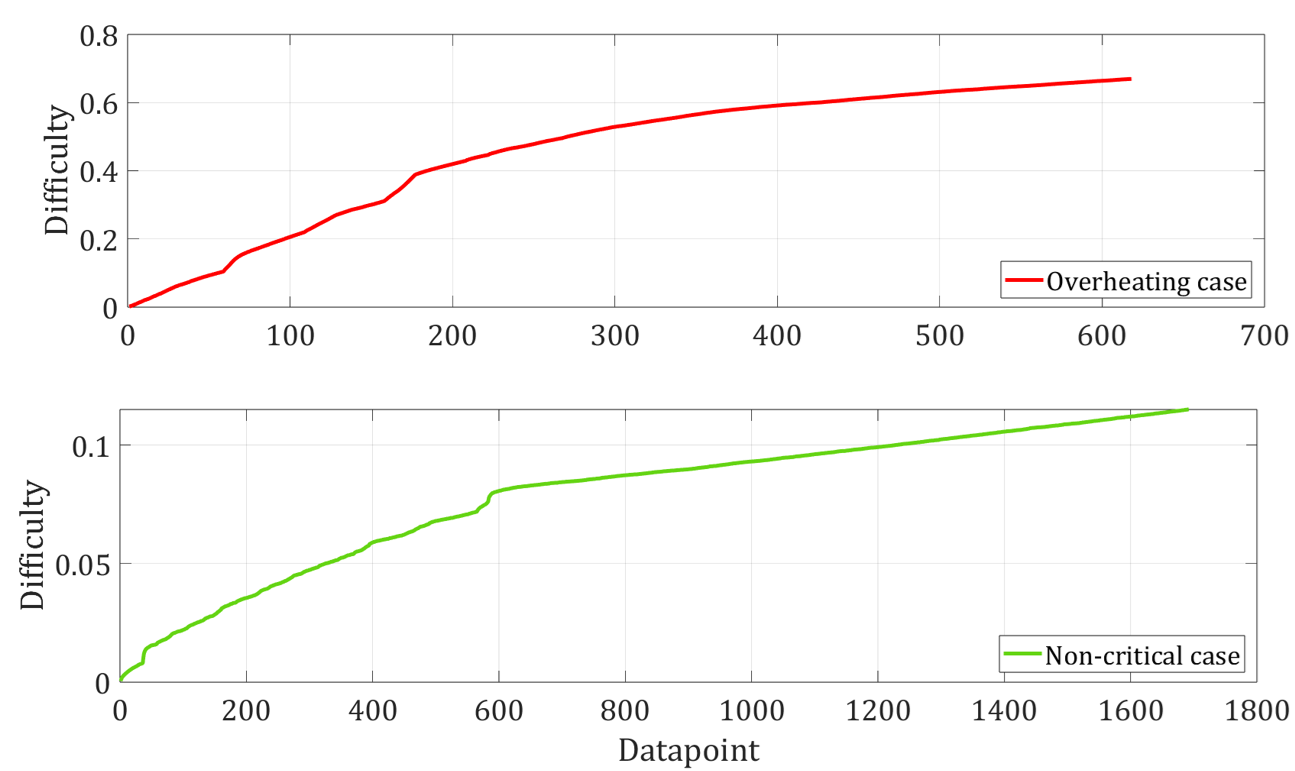}    
		\caption{Thermal difficulty for critical$\slash$non-critical data.} 
		\label{fig:difficulty}
	\end{center}
\end{figure}

 is the thermal difficulty of the $k$-th motor for the time horizon $t \in [t_l,t_h]$. $|e_{r_k}(t)|$ is the absolute value of the  current reconstruction error related to the targeted profile $x(t)$ of the   $k$-th joint state cumulatively for all channels (i.e., joint positions, velocities, and torques). According eq. \ref{TT} and Fig~\ref{fig:difficulty},  $d_k \rightarrow 1$ for anomalous $x(t)$. For thermally non-anomalous  $x(t)$,  $d_k\rightarrow 0$. Robots  can mutually exchange their $d$ to be aware in advance of the respective capability  to contribute to common goals without getting stuck in a thermal saturation.  Robots informed by their   VAE-embedded  DTw that provide $d$ can thermally anticipate their self-orchestration without human intervention, as expected {in Industry 6.0 ((\cite{carayannis2024toward}).} Although $|e_{r_k}(t)|$ indicates outliers, it was observed while training VAE-models that it does not necessarily decrease as the number of epochs increases (see table \ref{reconepoc}). This is also due to the random \mbox{initialization of neural weights.}


\begin{table}[h!] 
	\begin{center}
		\caption{Normalized  reconstruction error $r_e$ as a function of the number of epochs $n_e$.}
		\begin{tabular}{cccccc}	\label{reconepoc}
			Channel & $n_e=20$ & $n_e=40$ & $n_e=60$ & $n_e=120$ & $n_e=240$ \\\hline
			Torque & 0.1740 & 0.1163 & 0.1781  & 0.0832 & 0.2079 \\
			Position & 0.4132  &  0.3576 & 0.1621 & 0.1542 & 0.0885 \\ 
			Velocity & 0.7043 & 0.4967 & 0.5992 & 0.4395
			 & 0.3185 \\ \hline
		\end{tabular}
	\end{center}
\end{table}

\begin{figure}[b!]
	\begin{center}
		\includegraphics[width=1\columnwidth]{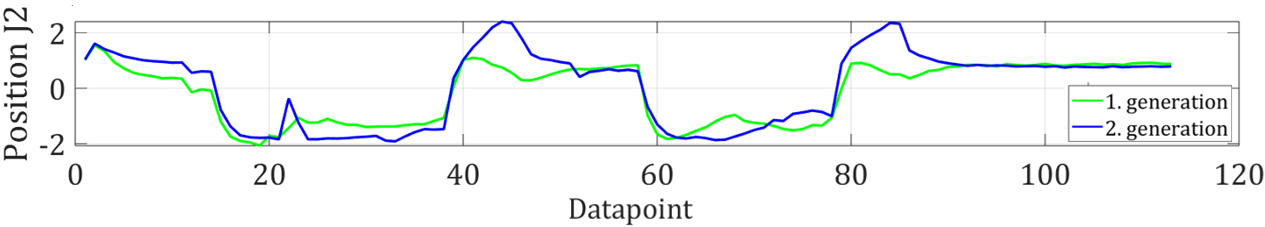}    
		\caption{Non-critical motions created by a VAE-based DTw.} 
		\label{fig:gen1}
	\end{center}
\end{figure}
\subsection{Self-achievement implications for Society 6.0}
Humans  use physical contact interaction forces to communicate and guide compliant collaborative robots. The desired symbiosis between humans and robots is conditioned by an engaging user perception of robots for self-fulfillment. To avoid threatening thermal burn hazards (see EN ISO 13732-1:2006), employ robots to augment humans, and support the comfortable autonomy   in societal tasks (e.g., household, healthcare, etc),  robot manufacturers automatically shut down the robot to meet their social responsibility in terms of safety (\cite{Kinova}, \cite{Franka}). Beyond shutdowns, a VAE-driven DTw  uses the predictive thermal difficulty presented thus far to convey a spatial insight into the missing thermal awareness of robot motors (see, e.g., upper l.h.s  of Fig.~\ref{fig:outlook}). In combination with augmented reality, the Gen AI-embedded DTw allows the  robot to predict the overheating behavior of its motors using spheres with a color gradient that maximizes the thermal salience in visualized postures accessible to almost everyone  and everywhere. The omnipresent and embodied AI  helps anticipate thermal burns and recover from a thermal saturation through manual posture optimization  and avoid downtime. This reconfiguration can also be achieved using the generative skills of the VAE-driven DTw. In fact, Fig.~\ref{fig:gen1} shows two uncritical motions created using different $\epsilon$ in eq. \ref{repara} while remaining on the learned data distribution. 

\section{Conclusion}

This work captures the distribution of joint data for the discriminative and generative thermal management of robot actuators.
It introduces the thermal difficulty score as a means to share the thermal condition of robot joints for a desired motion. The score depends upon the reconstruction error of the variational autoencoder with which the digital twin of the robot is equipped to detect thermally anomalous  and generate thermally uncritical state of joints. The underlying two dimensional  latent space is used to transform captured distributions associated with non-anomalous joint states into  opportunities for robotized application in the \mbox{society 6.0 and industry 6.0 realm.}%

\subsubsection{Acknowledgments:}
 {We would like to thank Mr. T. K. La for data collection.}

\bibliography{ifacconf}             
                                                   







\end{document}